\documentclass{article}

\usepackage[preprint]{neurips_2022}

\usepackage{natbib}
\setcitestyle{numbers,square}

\usepackage[utf8]{inputenc} 
\usepackage[T1]{fontenc}    
\usepackage{hyperref}       
\usepackage{url}            
\usepackage{booktabs}       
\usepackage{amsfonts}       
\usepackage{nicefrac}       
\usepackage{microtype}      
\usepackage{xcolor}         
\usepackage{amsmath}        
\usepackage{graphicx}       
\usepackage{subcaption}
\usepackage{algorithm}
\usepackage{bbm}
\usepackage{algpseudocode}
\usepackage{multirow}
\usepackage{tikz}
\usepackage{bbding}
\usepackage[most]{tcolorbox}

\title{AdvFunMatch: When Consistent Teaching Meets Adversarial Robustness}

\author{
  Zihui Wu \\
  Xidian University \\
  \texttt{zihui@stu.xidian.edu.cn} \\
  \And
  Haichang Gao \\
  Xidian University \\
  \texttt{hchgao@xidian.edu.cn} \\
  \And
  Bingqian Zhou \\
  Xidian University \\
  \texttt{zbqxidian@stu.xidian.edu.cn} \\
  \And
  Ping Wang \\
  Xidian University \\
  \texttt{pingwangyy@foxmail.com} \\
}
\begin{document}
\maketitle

\begin{abstract}

    \emph{Consistent teaching} is an effective paradigm for implementing knowledge distillation (KD), where both student and teacher models receive identical inputs, and KD is treated as a function matching task (FunMatch). However, one limitation of FunMatch is that it does not account for the transfer of adversarial robustness, a model's resistance to adversarial attacks. To tackle this problem, we propose a simple but effective strategy called Adversarial Function Matching (AdvFunMatch), which aims to match distributions for all data points within the $\ell_p$-norm ball of the training data, in accordance with consistent teaching. Formulated as a min-max optimization problem, AdvFunMatch identifies the worst-case instances that maximizes the KL-divergence between teacher and student model outputs, which we refer to as "mismatched examples," and then matches the outputs on these mismatched examples. Our experimental results show that AdvFunMatch effectively produces student models with both high clean accuracy and robustness. Furthermore, we reveal that strong data augmentations (\emph{e.g.}, AutoAugment) are beneficial in AdvFunMatch, whereas prior works have found them less effective in adversarial training. Code is available at \url{https://gitee.com/zihui998/adv-fun-match}.

\end{abstract}
\section{Introduction}
Deep neural networks have achieved state-of-the-art (SOTA) performance in various domains, including computer vision, natural language processing, and speech recognition. However, the considerable size and computational complexity of these models limit their deployment on resource-constrained devices. Knowledge distillation (KD) \cite{cite9} techniques offer a solution to this issue by enabling knowledge transfer from a larger teacher model to a smaller, more efficient student model. While KD traditionally aims to "distill" a teacher model, the recent work FunMatch \cite{cons_t} reinterprets KD as a task of matching the functions implemented by the teacher and student. It subsequently introduces the \emph{consistent teaching} principle, which asserts that both student and teacher models should receive identical inputs during the training process. Consequently, FunMatch can prevent overfitting even over a large number of training epochs, achieving remarkable results on image classification tasks.

Nevertheless, while KD has effectively transferred general knowledge, transferring adversarial robustness—the ability of a model to resist against adversarial perturbations—has proven challenging \cite{cite3}. Previous research \cite{ard,akd,iad,rslad} has investigated robustness in the context of KD and shown that distilling on adversarial examples \cite{Goodfellow} can successfully transfer adversarial robustness. However, these studies have not explicitly focused on the consistent teaching principle, which has demonstrated its effectiveness in implementing KD for non-robust image classification tasks. This motivates the current paper: 

\begin{center}
\emph{Can robustness transfer be improved in KD when adhering to the consistent teaching principle?}
\end{center}

In response to this question, we introduce Adversarial Function Matching (AdvFunMatch), a novel method that leverages the principles of consistent teaching to enhance the transfer of adversarial robustness from the robust teacher model to the student model. Instead of merely matching on clean examples as in FunMatch, AdvFunMatch aims to match distributions for each data point within the $\ell_p$-norm ball of clean examples. To achieve this goal, we formulate AdvFunMatch as a minimax optimization problem with the objective of identifying worst-case instances that maximize the KL-divergence between teacher and student model outputs. These instances are termed "mismatched examples" because, in contrast to traditional adversarial examples that focus on maximizing the divergence from the one-hot class label, our approach targets instances that yield the highest discrepancy between teacher and student model outputs. By matching on the worst-case instances with the highest discrepancy, we can indirectly align distributions of other data points in the norm ball, thereby accomplishing the initial goal.

To generate these mismatched examples, we utilize the Projected Gradient Descent (PGD) technique, as previously employed in \cite{pgd_at}. PGD leverages the signed gradient flow to create input perturbations. As our AdvFunMatch aims to match the distribution throughout the entire norm ball, we integrate the teacher model's gradient into the perturbation update. It offers a more accurate representation of the target distribution's variation trend and thus contributes to generating more effective mismatched examples that lead to a larger discrepancy between the teacher and student models.
 Guided by the teacher gradient flow, we demonstrate that even a 2-step PGD can produce examples with sufficiently high discrepancy, making the training of AdvFunMatch more efficient.

Our experimental results demonstrate that AdvFunMatch effectively generates student models with high clean and robust accuracy, outperforming previous adversarial training and robustness distillation methods. Furthermore, we find that strong data augmentation techniques, such as AutoAugment \cite{autoaugment} and MixUp \cite{mixup}, are beneficial within the AdvFunMatch framework, in contrast to previous works that reported limited effectiveness in adversarial training \cite{cite8,awp}. This observation suggests that the restricted impact of strong data augmentation on robustness might be due to inappropriate label settings.

The main contributions of our work can be summarized as follows.

\begin{itemize}
\item We propose AdvFunMatch, a simple but effective robustness distillation framework that leverages the principles of consistent teaching to transfer adversarial robustness from a teacher model to a student model.
\item We introduce the concept of "mismatched examples" to identify worst-case instances that maximize the discrepancy between teacher and student model outputs. 
\item We demonstrate the effectiveness of AdvFunMatch through extensive experiments on benchmark datasets, showing that it outperforms traditional adversarial training and robustness distillation methods in terms of both clean accuracy and adversarial robustness.  For example, on the CIFAR-10 dataset, AdvFunMatch resulted in a ResNet-18 student model with a clean accuracy of 89.08\% and an AutoAttack \cite{autoattack} accuracy of 58.30\%,
surpassing previous SOTA method \cite{sehwag2021robust} by 4.49\% and 2.76\%, respectively.

\item We find that strong data augmentations, such as AutoAugment, can significantly improve the performance of AdvFunMatch, suggesting that the restricted impact of strong data augmentation on robustness might stem from inappropriate label settings.
\end{itemize}

The remainder of this paper is organized as follows: Section \ref{sec:2} provides background information on knowledge distillation, consistent teaching, adversarial robustness and data augmentation; Section \ref{sec:3} presents our proposed AdvFunMatch method; Section \ref{sec:4} outlines the experimental setup and discusses results; and finally, Section \ref{sec:5} concludes the paper and highlights potential avenues for future research.

\section{Background and Related Work}
\label{sec:2}
In this section, we discuss prior work related to our study, which encompasses knowledge distillation, consistent teaching, adversarial robustness, and data augmentation techniques.

\begin{figure*}[ht]
        \centering
        \includegraphics[width=\linewidth]{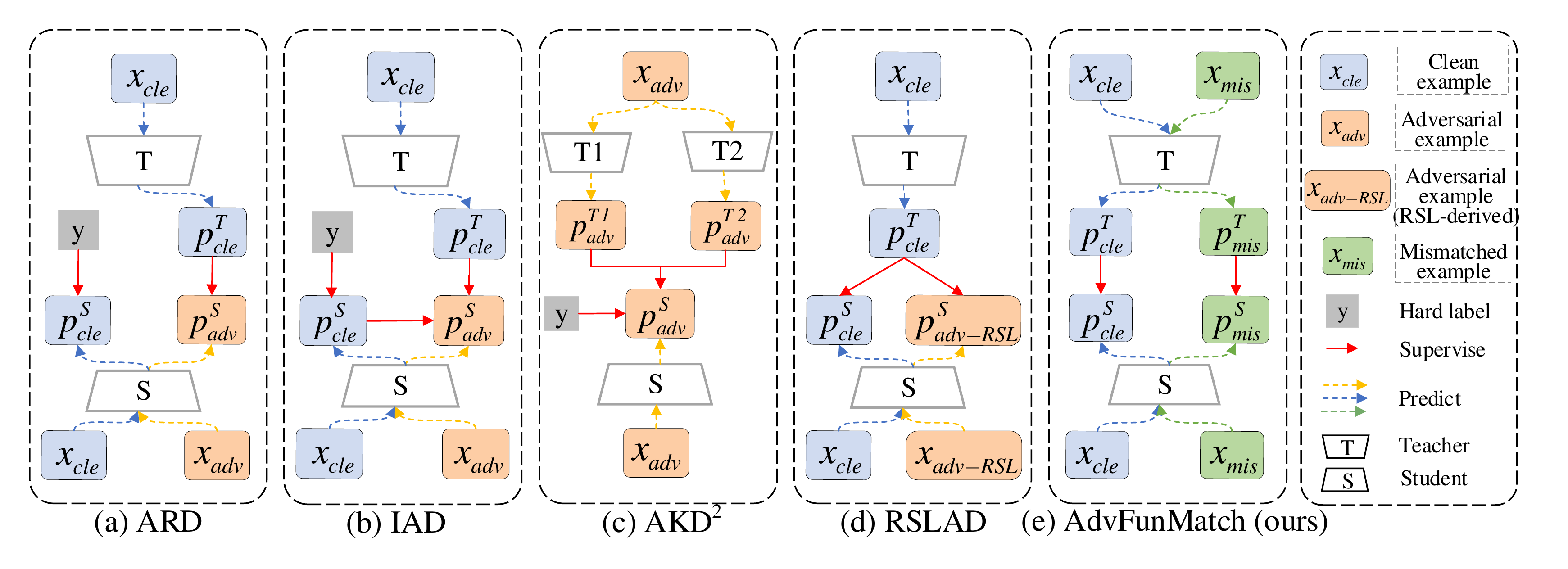}
        \caption{ An illustration of different approaches for adversarial robustness distillation. The arrows represent the input and output relations of the teacher and student models. The labels denote the kinds of examples employed for matching. The proposed approach (AdvFunMatch) leverages consistent teaching and mismatched examples to facilitate the transfer of adversarial robustness from the teacher model to the student model.}
    \label{fig:1}
\end{figure*}

\textbf{Knowledge Distillation}.
KD is a technique designed to transfer knowledge from a larger, more complex teacher model to a smaller, more computationally efficient student model. Hinton et al. \cite{cite9} first introduced KD as a model compression method, where a student model is trained to mimic the softened output distribution of a teacher model. Since then, various improvements have been proposed, including attention-based distillation \cite{attention_}, feature-based distillation \cite{gift_,hint_}, and relational distillation \cite{rkd_}. Despite the success in transferring general knowledge, prior work has demonstrated that standard KD struggles to transfer adversarial robustness \cite{cite3}.

\textbf{Consistent Teaching}.
Consistent teaching is a learning paradigm in which both the student and teacher models receive identical inputs during the distillation process. The concept of consistent teaching is exemplified by the FunMatch method \cite{cons_t}, which treats the student and teacher models as two functions and matches their outputs with identical inputs (including augmentations). It is worth mentioning that FunMatch does not use hard labels as supervisory signals, and our AdvFunMatch follows this setting. 
Let $T(x)$ denote the output of the teacher model $T$ and $S(x)$ denote the output of the student model $S$. Mathematically, the FunMatch task can be formulated as:

\begin{equation}
\min_{\theta_s} \mathbb{E}_{x \in \mathcal{D}^\prime }  KL(T(x) \Vert S(x)),
\end{equation}

where $\theta_s$ represents the parameters of the student model $S$, $KL$ represents the KL-divergence,
and $\mathcal{D}^\prime$ denotes the augmented dataset.

\textbf{Adversarial Robustness}.
Deep neural networks are susceptible to adversarial attacks, as highlighted by previous studies \cite{Goodfellow,Szegedy}. To mitigate this issue, various methods have been proposed, with adversarial training (AT) \cite{pgd_at,trades,mart} being the most effective one. However, AT requires larger model capacity than standard training, which limits its application on resource-limited devices. To address this problem, researchers have explored using model pruning \cite{YeLX0CLZZMW19,Sehwag0MJ20,zhaoholistic,dnr} and KD \cite{cite2,ard,akd,iad,rslad,ChanTO20,Shao,Maroto} to reduce the capacity of robust models while maintaining high robustness. In this paper, we focus on the KD approach and find that prior robustness distillation studies have not explicitly focused on consistent teaching, which our proposed AdvFunMatch framework addresses.

For instance, Figure \ref{fig:1} shows that previous methods such as ARD \cite{ard}, IAD \cite{iad}, and RSLAD \cite{rslad} all match adversarial example outputs of the student model with clean example outputs of the teacher model, which violates the rule of identical inputs. AKD$^2$ \cite{akd} satisfies consistent teaching by matching adversarial examples with two teacher models, a standard teacher, and a robust teacher. However, we find that matching on traditional adversarial examples (i.e., the worst-case instances furthest from the hard labels) is not the best approach for our AdvFunMatch method. This insight leads us to propose the use of mismatched examples. It is worth noting that RSLAD does not use hard label adversarial examples but instead employs "robust soft label" (RSL), which refers to outputs of the robust teacher model on clean examples. Nevertheless, we show that RSL-derived adversarial examples also fail to identify the worst-case instances that maximize the mismatch between teacher and student models.

\textbf{Data Augmentation}.
Data augmentation techniques, such as random cropping, flipping, and color jittering, have been widely used to enhance the generalization capabilities of deep learning models. However, previous works \cite{cite8,awp} have found limited effectiveness of stronger data augmentation techniques, such as AutoAugment \cite{autoaugment}, RandAugment \cite{randaugment}, MixUp \cite{mixup}, Cutout \cite{cutout}, and CutMix \cite{cutmix}, in AT when compared to simpler techniques such as flip and random crop. Previous work has explored how to effectively utilize these strong data augmentations in AT, including combining with weight averaging (WA) \cite{wa} regularization \cite{RebuffiGCSWM21} and using dual batch normalization structure \cite{dual_bn}. A recent work \cite{da_alone} proposed a new augmentation method customized for AT that balances hardness and diversity.

In contrast, Beyer et al. \cite{cons_t} suggest that strong data augmentations are beneficial for the FunMatch task, and Wang et al. \cite{wang2022makes} studied the effects of various strong data augmentation techniques in KD and showed they can effectively improve the standard classification accuracy. The discrepancy of data augmentations in KD and AT inspired us to investigate their role in AdvFunMatch.

Our work shows that all the above-mentioned strong data augmentation techniques can significantly improve robust generalization within the AdvFunMatch framework. Additionally, we confirm the effectiveness of data augmentations in previous robustness distillation methods, such as RSLAD and ARD, although not as effective as in AdvFunMatch. This suggests that the limited impact of strong data augmentation on AT may be due to inappropriate label settings.

In summary, our work builds upon the foundations laid by research in KD, consistent teaching, adversarial robustness, and data augmentation techniques. We propose AdvFunMatch, a novel method that leverages consistent teaching principles to effectively transfer adversarial robustness from a teacher model to a student model while maintaining high clean accuracy.
\section{Proposed Method}
\label{sec:3}
In this section, we present our proposed method, AdvFunMatch, which aims to transfer adversarial robustness from a teacher model to a student model while ensuring consistent teaching. In section 3.1, we introduce the overall learning objective of AdvFunMatch. In section 3.2, we delve into the optimization details, including why we formulate AdvFunMatch as a minimax optimization problem that focuses on matching the mismatched examples, and we describe how to generate these examples.

\subsection{Problem Formulation}
\textbf{AdvFunMatch}. Although FunMatch provides a good implementation for KD that improves the student model's generalization, it also shares the limitations of KD in failing to transfer adversarial robustness. To address this issue, we propose AdvFunMatch. Instead of solely focusing on the augmented samples, AdvFunMatch requires the student model to match every data point within the $\ell_p$-norm ball of the augmented sample, emphasizing the importance of robustness in the process. This is formulated in Eq. \ref{eq:2}.

\begin{equation}
\label{eq:2}
\min_{\theta_s} \mathbb{E}_{x \in \mathcal{D}^\prime}KL(T(x^\prime) \Vert S({x^\prime})), \forall x^\prime \in \mathcal{B}(x,\epsilon)
\end{equation}

While input perturbations can also be regarded as a type of data augmentation, we utilize $\mathcal{B} (x, \epsilon )$ to represent the norm ball with a radius of $\epsilon$ on the samples in the augmented dataset $\mathcal{D}^\prime$, differentiating it from
conventional data augmentation techniques to emphasize the focus on adversarial robustness.

Despite the simplicity, we would like to point out that the learning objective of Eq. \ref{eq:2} significantly differs from previous works on AT and robustness distillation. Earlier efforts typically aimed to fit the worst-case instances within the norm ball that are farthest from the hard labels (i.e., adversarial examples) to a fixed distribution (e.g., hard labels or the teacher model's output distribution on clean samples), thereby endowing the model with adversarial robustness.
However, in AdvFunMatch, we aim to adhere to the principles of consistent teaching by requiring the student model to match the teacher model's outputs for every point within the norm ball of the training data. This implies that our target distribution is no longer a fixed probability distribution, but rather the teacher model's distribution throughout the entire norm ball. As we will discuss, the difference makes the optimization process of AdvFunMatch slightly more complex than previous methods.

\subsection{How to Optimize AdvFunMatch?}
\begin{figure*}[t]
\vskip 0.2in
\center{
    \begin{subfigure}{0.32\linewidth}
        \centering
        \includegraphics[width=0.9\linewidth]{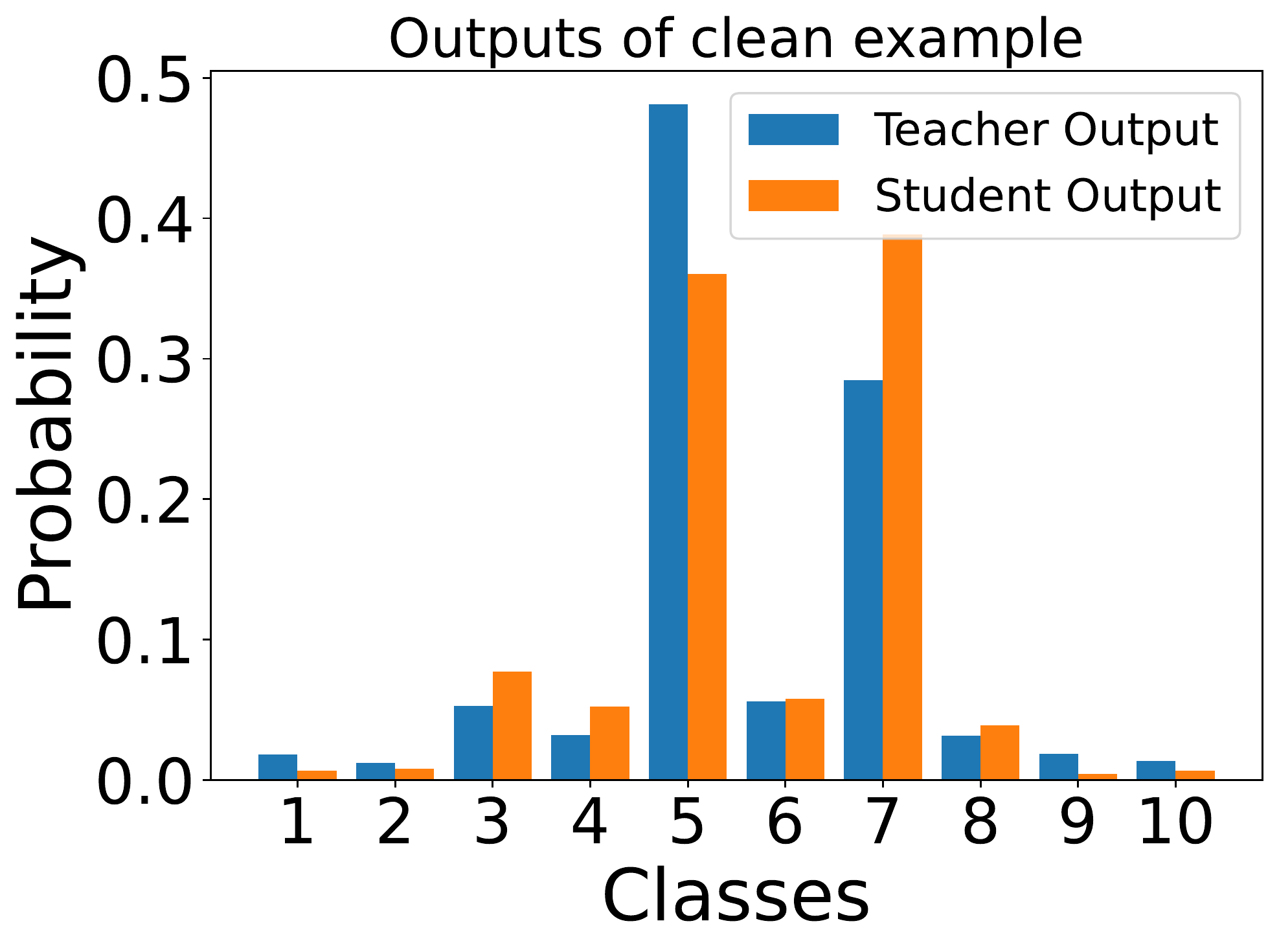}
        \caption{Clean example} 
        \label{fig:1a} 
    \end{subfigure}
    \begin{subfigure}{0.32\linewidth}
        \centering
        \includegraphics[width=0.9\linewidth]{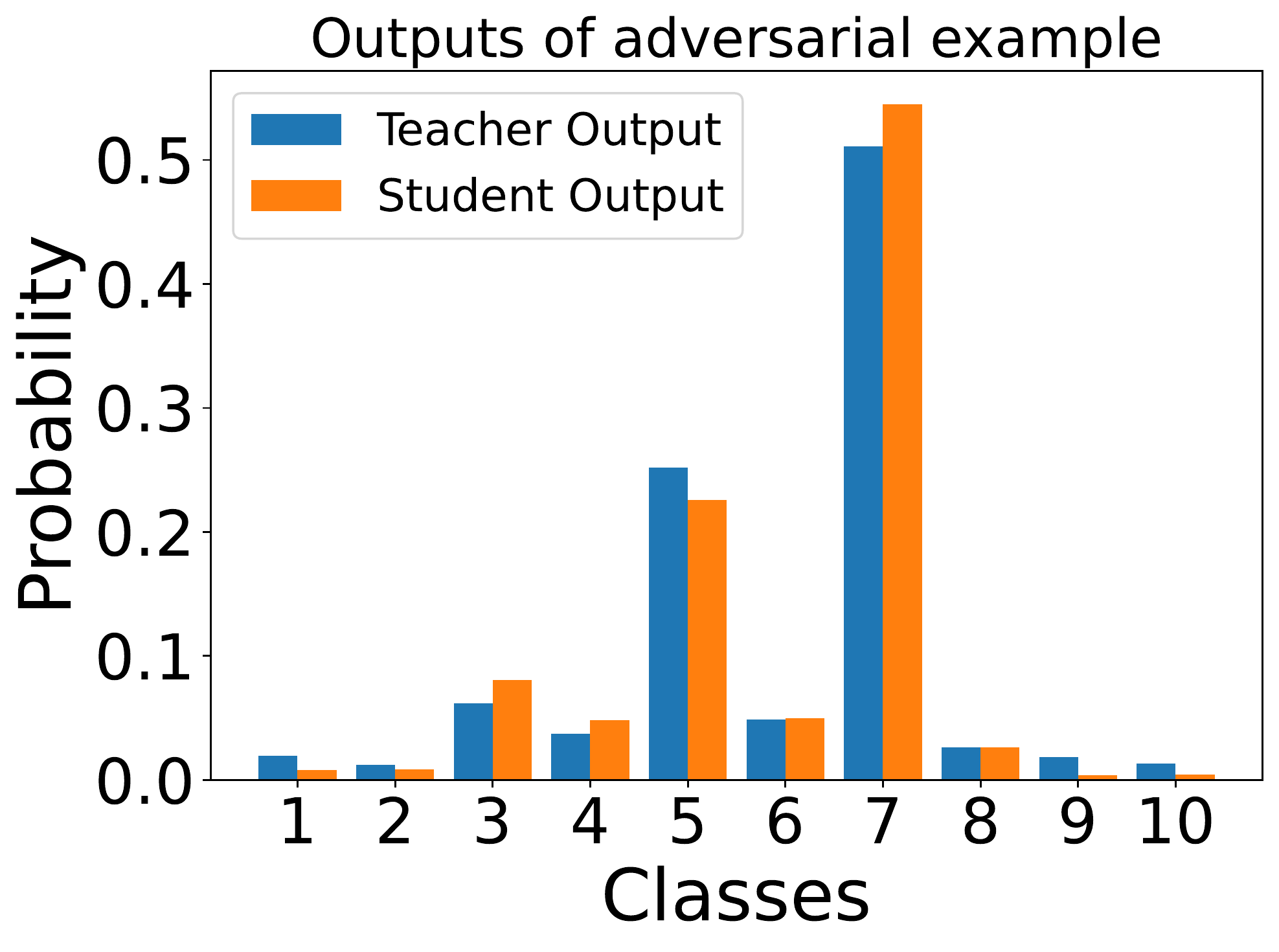}
        \caption{Adversarial example}
        \label{fig:1c}
    \end{subfigure}
    \begin{subfigure}{0.32\linewidth}
        \centering
        \includegraphics[width=0.9\linewidth]{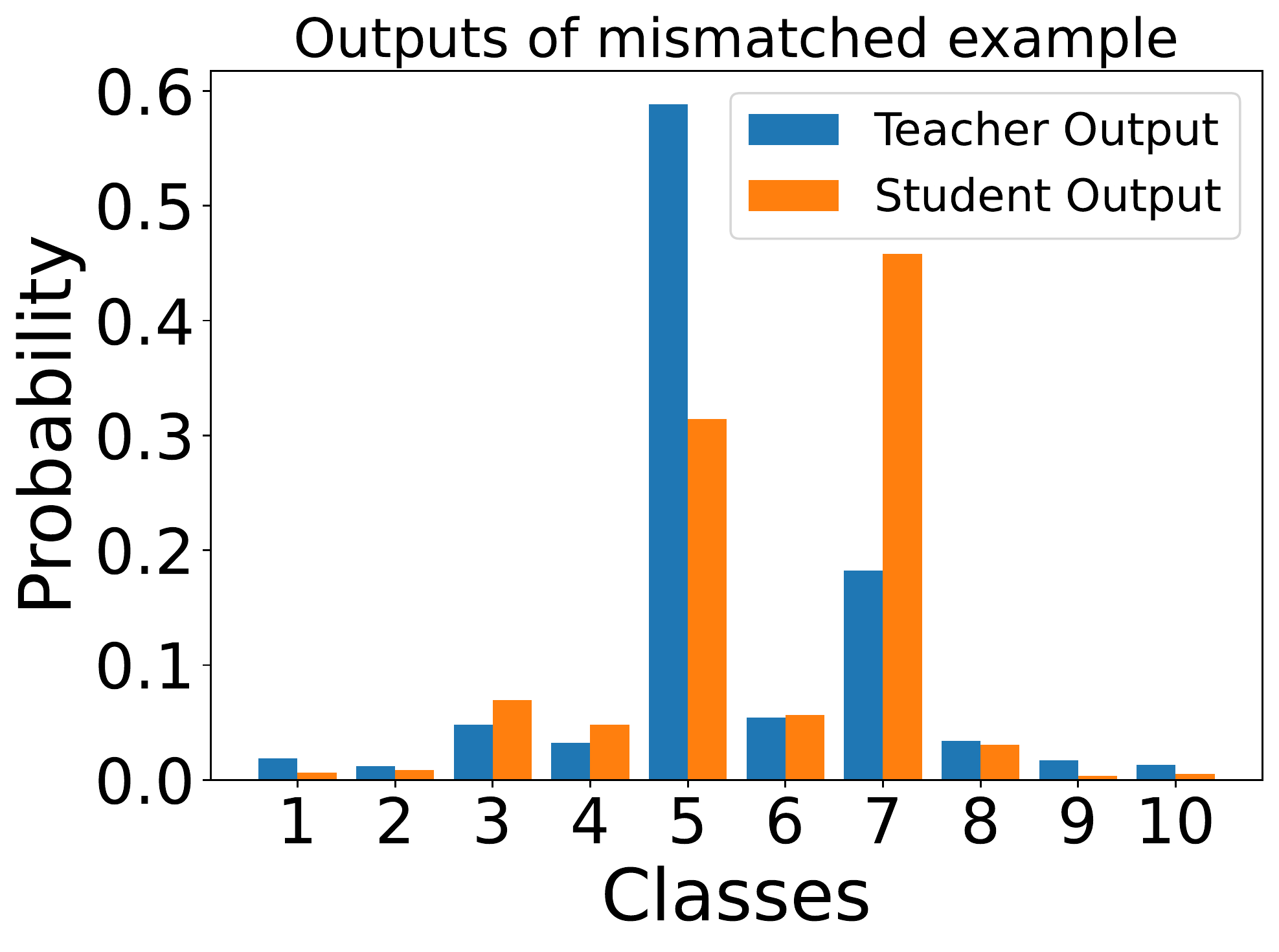}
        \caption{Mismatched example}
        \label{fig:1d}
    \end{subfigure}
}
    \caption{Output probability distributions of teacher and student models on clean example, adversarial example and mismatched example. Adversarial example is generated by using the student model.} 
\label{fig:2}
\vskip -0.2in 
\end{figure*}

\textbf{Mismatched example}. 
Due to the continuous nature of the input space, sampling all points within the norm ball to solve Eq. \ref{eq:2} is infeasible. To solve this problem, we follow prior works in AT and convert Eq. \ref{eq:2} into a minimax optimization problem, as shown in Eq. \ref{eq:3} and Eq. \ref{eq:4}. In Eq. \ref{eq:3}, we seek the worst-case instance $\tilde{x}$ that maximizes the discrepancy between the teacher and student model outputs. As this objective is quite different from that of previous adversarial examples, we refer to these worst-case instances as "mismatched example" for clarity.

In fact, we discover that adversarial examples do not always correspond to the points where teacher and student models exhibit the highest discrepancy. Figure \ref{fig:2} effectively illustrates this phenomenon by depicting the output probability distributions for various example types. It is noticeable that the adversarial example prompts both the teacher and student models' output distributions to deviate from the ground truth class "5"; however, these distributions still largely align. In contrast, the mismatched example specifically targets the maximization of discrepancies between the teacher and student models.

\begin{equation}
\label{eq:3}
\tilde{x} \leftarrow \arg\max\limits_{\tilde{x} \in \mathcal{B} (x, \epsilon )} KL(T(\tilde{x}) \Vert S(\tilde{x}))
\end{equation}

\begin{equation}
\label{eq:4}
\min_{\theta_s} \mathbb{E}_{x \in \mathcal{D}^\prime } (1-\lambda) KL(T(x) \Vert S(x)) + \lambda KL(T(\tilde{x}) \Vert S(\tilde{x})) 
\end{equation}

We match the mismatched examples using Eq. \ref{eq:4}, which balances the robustness-accuracy tradeoff by also including the original FunMatch term, i.e., the distillation loss on clean samples $x$. Intuitively, by matching the worst-case point with the largest discrepancy, we can indirectly match other points within the norm ball, thereby solving Eq. \ref{eq:2}.

\textbf{Generating mismatched example through teacher gradient guidance}.
Consistent with previous work in AT \cite{pgd_at,trades}, we employ the PGD method to identify the worst-case instances within the norm ball.
PGD employs the signed gradient flow to the inputs as perturbation, and iteratively updates to generate the worst-case examples, as formulated in Eq. \eqref{Eq:extra}.
\begin{equation}
\label{Eq:extra}
\tilde{x} \leftarrow \Pi_{\mathcal{B}(x,\epsilon)}(\tilde{x}+\eta \cdot sign(\nabla_{\tilde{x}} \mathcal{L}_{max})),
\end{equation}
where $\Pi$ is the clip operator, $\eta$ is the step size and
$\mathcal{L}_{max}$ is the loss function aimed to maximize.

\begin{figure}[t]
  \centering
  \begin{minipage}[b]{0.49\textwidth}
    \includegraphics[width=\textwidth]{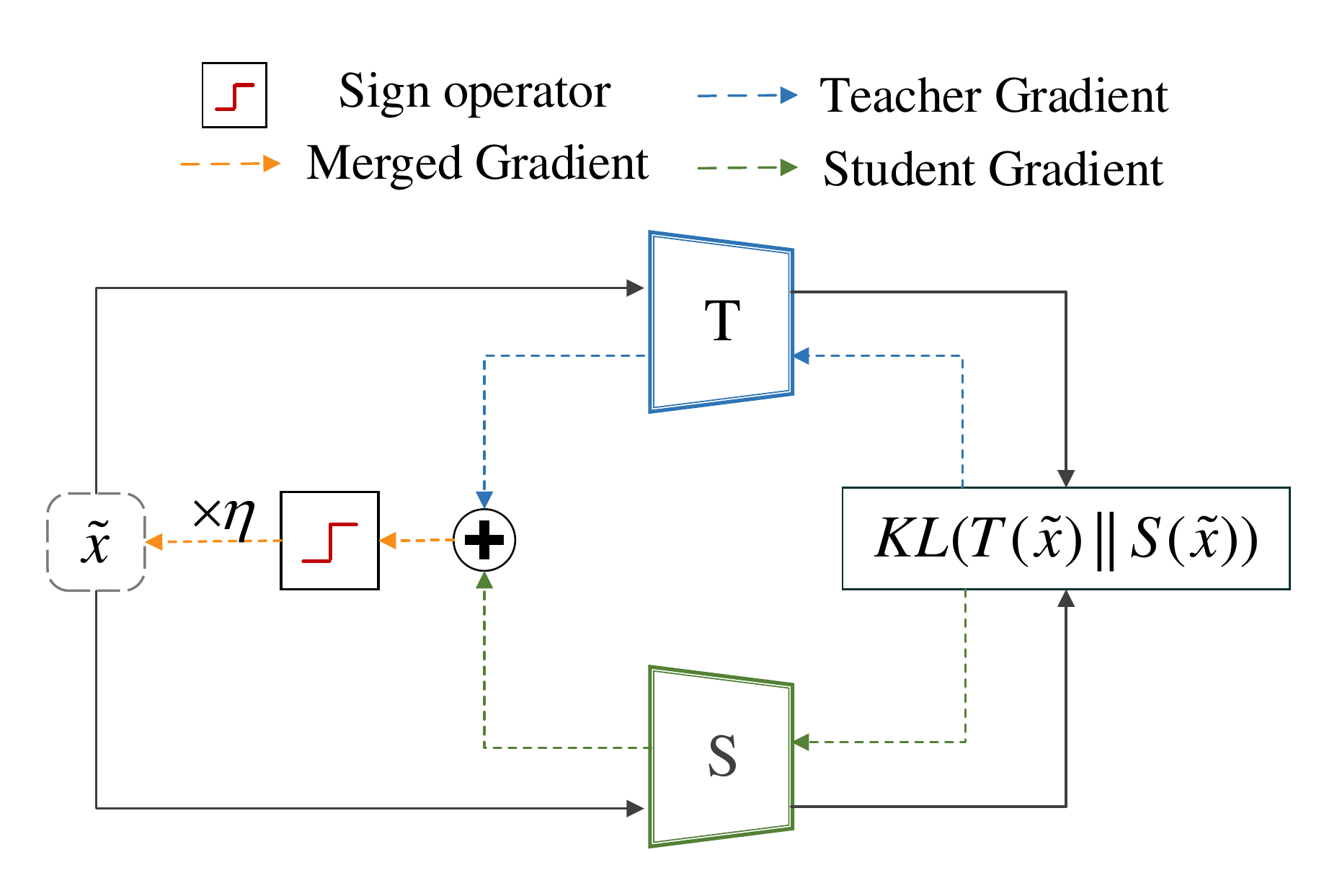}
    \caption{Generate the mismatched example $\tilde{x}$ via the guidance of teacher gradients, $\eta$ is the step size.}
    \label{fig:3}
  \end{minipage}
  \hfill
  \begin{minipage}[b]{0.49\textwidth}
    \includegraphics[width=\textwidth]{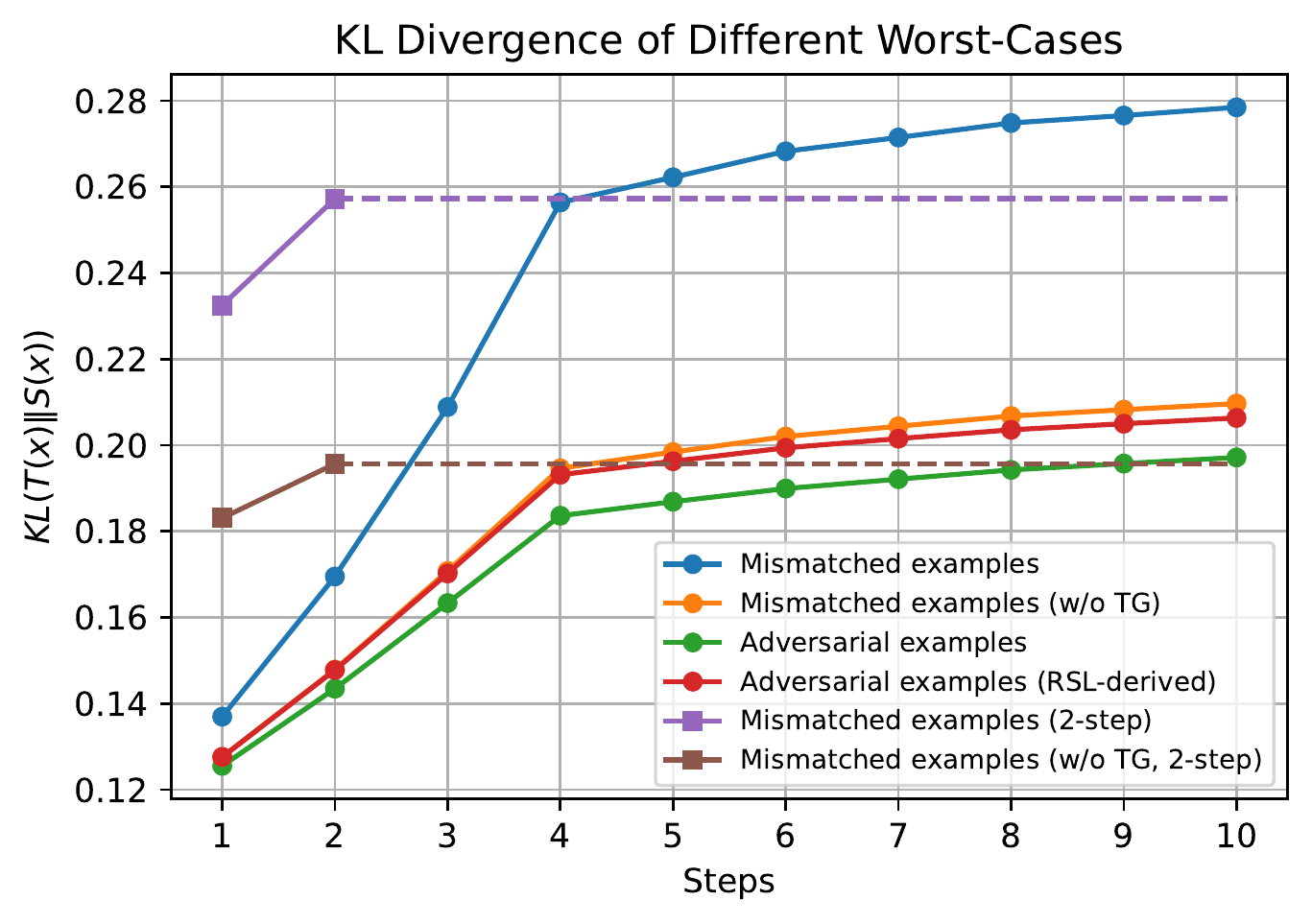}
    \caption{KL-divergence between teacher and student model's output for different examples, TG means teacher gradient.}
    \label{fig:4}
  \end{minipage}
\end{figure}

In our AdvFunMatch, we substitute $\mathcal{L}_{max}$ with the KL-divergence term in Eq. \ref{eq:3} to generate mismatched examples that maximize the discrepancies between the teacher and student models. Interestingly, we find that a specific implementation detail plays a crucial role in optimizing mismatched examples: whether to allow backpropagation through the teacher model. More specifically, in PyTorch, this is reflected in whether to detach the logits output of the teacher model. To provide a better understanding, we visualize this process in Figure \ref{fig:3}.

In Figure \ref{fig:4}, we compare the optimization effects with and without the involvement of the teacher's gradient, illustrating the differences between the two approaches. It becomes evident that when the teacher's gradient flow contributes to the generation of mismatched examples, the produced samples substantially intensify the discrepancies between the models.
The reason for this difference is that in AdvFunMatch, our target distribution is no longer a fixed distribution, but rather the teacher model's probability distribution within the norm ball. This means that during the PGD iteration process, the target distribution changes as the samples update. Therefore, using the teacher model's gradient helps describe the trends in the target distribution's changes, which in turn guides the generation of mismatched examples.

It is worth mentioning that earlier works \cite{ChanTO20,Shao} have also utilized the teacher model's gradient to train student models with the objective of improving their robustness. However, their approach focused on aligning the gradients of the student and teacher models at the input level, whereas our method leverages the teacher model's gradient to generate more mismatched samples. This key difference in the utilization of the teacher's gradient sets our approach apart from the earlier works.

\textbf{Achieving Robustness with 2-Step PGD}. Prior research on AT and robustness distillation typically employs 10-step PGD adversarial examples during training. However, computing both student and teacher gradients for 10 steps to generate mismatched examples proves to be highly computationally demanding. Earlier efforts to enhance computational efficiency utilized 1-step Fast Gradient Sign Method (FGSM) \cite{Goodfellow} adversaries for training. However, FGSM-based AT tends to be unstable and frequently leads to catastrophic overfitting \cite{gradalign}.

In this paper, we use a 2-step PGD approach for generating mismatched examples. To ensure faster convergence, we use a large step size $8/255$ rather than
the commonly used $2/255$ in the 10-step PGD. Our experiments demonstrate that with just 2 steps, the mismatched examples generated using the teacher's gradient guidance outperform commonly used 10-step adversarial examples and 10-step RSL-derived adversarial examples in increasing the output discrepancy between the student and teacher models, as illustrated in Figure \ref{fig:4}. 

\textbf{Aggressive $\epsilon$ Strategy}. To further enhance robust generalization, we propose an aggressive $\epsilon$ strategy: before generating each mismatched example, we sample a random perturbation size $\epsilon_{r}$ from the interval [0, $\epsilon_{add}$], where $\epsilon_{add}$ denotes the maximum additional perturbation size. We then use $\epsilon_{train} + \epsilon_{r}$ as the maximum perturbation size for the current sample. This simple strategy allows the student model to match the teacher under larger norm constraints, and compared to directly increasing the training perturbation size, the randomly sampled perturbation size
adds to the diversity of the samples.
In Section \ref{sec:4.3}, we show that this strategy further boosts the model's robustness, albeit at the cost of a small reduction in clean accuracy. Since our primary focus in this paper is on robustness, we employ this approach to generate mismatched examples in all subsequent experiments.

\section{Experiments}
\label{sec:4}

\subsection{Experimental Setup}

\textbf{Datasets and model architectures}.
Our experiments are conducted on CIFAR-10 and CIFAR-100 \cite{krizhevsky2009learning} datasets.
We examine two student networks, specifically ResNet-18 \cite{resnet} and MobileNetV2 \cite{mnv2}. For ResNet-18, we employ the WideResNet-28-10 (WRN-28-10) \cite{wrn} provided by Wang et al. \cite{wang} as the teacher model for both the CIFAR-10 and CIFAR-100 datasets. These two models are trained using an extensive amount of synthetic data and undergo numerous training epochs, ultimately achieving SOTA performance in terms of both clean accuracy and adversarial robustness, as illustrated in Table \ref{tab:1}. For the lightweight MobileNetV2, we utilize our ResNet-18 models in Table \ref{tab:6} as the teacher.

\textbf{Training procedure}.
We train the student models using the SGD optimizer with Nesterov momentum. The momentum factor is set to 0.9, and the weight decay is set to 2 × 10$^{-4}$. We employ a cosine learning rate schedule with a 5-epoch warmup, and the batch size is set to 512. After the warmup, the initial learning rate is 0.4. To maintain consistency with prior robustness distillation method \cite{rslad}, we train for 300 epochs. However, we also investigate the impact of longer epochs and demonstrate that they improve performance under AdvFunMatch without causing robust overfitting \cite{cite8}. Unless mentioned otherwise, we utilize the commonly used Flip+RandomCrop as the data augmentation method. For all experiments, we set the temperature parameter to 1, and the balance parameter $\lambda$ in Eq. 4 is set to 0.9. 
We use the 2-step mismatched examples for training, the base training perturbation size $\epsilon_{train}$ is set to 8/255 under the $\ell_{\infty}$ norm. Moreover, based on our experiments presented in Table \ref{tab:4}, we choose the additional perturbation size $\epsilon_{add}$ = 6/255 as the default setting to achieve a better trade-off between robustness and accuracy in subsequent experiments.

\textbf{Evaluation metrics}. We evaluate the performance of our method using several metrics, including standard classification accuracy and robust accuracy under various attack methods, such as FGSM \cite{Goodfellow}, PGD \cite{pgd_at}, CW${\infty}$ \cite{cite3}, and AutoAttack (AA) \cite{autoattack}, with a maximum perturbation of $\epsilon_{test}$ = 8/255 for both CIFAR-10 and CIFAR-100 datasets.

\begin{table}[t]
\centering
\caption{Configurations and test accuracy (\%) under various attacks of the WRN-28-10 teacher models on CIFAR-10/CIFAR-100.}
\label{tab:1}
\resizebox{0.9\textwidth}{!}{%
\begin{tabular}{lcccccccc}
\toprule
Dataset   & Model  & Generated data & Training epochs & Clean & FGSM  & PGD20 & CW${\infty}$ & AA    \\ \midrule
CIFAR-10  & WRN-28-10 & 20M            & 2400            & 92.44   & 75.78 & 70.67 & 68.31        & 67.31 \\
CIFAR-100 & WRN-28-10 & 50M            & 1600            & 72.58   & 47.01 & 44.19 & 39.67        & 38.83 \\ \bottomrule
\end{tabular}%
}
\end{table}

\begin{table}[]
\centering
\caption{Test accuracy (\%) for various methods on CIFAR-10/CIFAR-100. Best result is in \textbf{bold}.}
\label{tab:2}
\resizebox{0.9\columnwidth}{!}{%
\begin{tabular}{lll|ccccc|ccccc}
\toprule
\multirow{2}{*}{Teacher}   & \multirow{2}{*}{Student}     & \multirow{2}{*}{Method} & \multicolumn{5}{c|}{CIFAR-10}                                                      & \multicolumn{5}{c}{CIFAR-100}                                                      \\
                           &                              &                         & Clean          & FGSM           & PGD20          & CW${\infty}$   & AA             & Clean          & FGSM           & PGD20          & CW${\infty}$   & AA             \\ \midrule
\multirow{7}{*}{WRN-28-10} & \multirow{7}{*}{ResNet-18}   & PGD-AT                  & 81.58          & 55.70          & 49.96          & 48.60          & 46.25          & 56.61          & 31.41          & 28.10          & 26.21          & 23.77          \\
                           &                              & TRADES                  & 82.35          & 57.98          & 51.49          & 49.85          & 48.35          & 56.20          & 31.51          & 29.25          & 25.40          & 24.27          \\
                           &                              & ARD                     & 83.37          & 58.92          & 52.80          & 51.33          & 49.18          & 60.98          & 36.05          & 32.87          & 29.24          & 27.08          \\
                           &                              & IAD                     & 83.66          & 58.81          & 52.59          & 50.67          & 48.40          & 57.80          & 34.10          & 31.37          & 27.55          & 25.78          \\
                           &                              & AKD$^2$                 & 84.30          & 59.24          & 52.22          & 51.15          & 48.86          & 60.16          & 33.01          & 29.72          & 26.95          & 25.03          \\
                           &                              & RSLAD                   & 84.91          & 59.59          & 52.89          & 51.04          & 49.03          & 60.78          & 36.09          & 32.49          & 28.61          & 26.97          \\
                           &                              & AdvFunMatch             & \textbf{86.58} & \textbf{64.28} & \textbf{55.94} & \textbf{54.80} & \textbf{52.40} & \textbf{64.30} & \textbf{38.46} & \textbf{34.29} & \textbf{30.65} & \textbf{28.80} \\ \midrule
\multirow{7}{*}{ResNet-18} & \multirow{7}{*}{MobileNetV2} & PGD-AT                  & 78.37          & 52.15          & 48.12          & 45.05          & 42.47          & 54.93          & 30.22          & 24.87          & 26.95          & 22.29          \\
                           &                              & TRADES                  & 78.08          & 53.76          & 48.88          & 45.97          & 45.09          & 55.34          & 30.09          & 27.43          & 22.95          & 21.85          \\
                           &                              & ARD                     & 81.56          & 58.05          & 52.32          & 50.45          & 48.69          & 59.95          & 36.02          & 32.99          & 29.29          & 27.56          \\
                           &                              & IAD                     & 81.69          & 57.93          & 53.12          & 50,13          & 47.96          & 55.50          & 33.70          & 31.55          & 27.78          & 26.20          \\
                           &                              & AKD$^2$                 & 83.33          & 57.77          & 52.51          & 50.45          & 48.30          & 59.90          & 33.30          & 30.23          & 27.27          & 25.24          \\
                           &                              & RSLAD                   & 84.31          & 59.45          & 53,36          & 51.24          & 49.12          & 60.16          & 35.47          & 32.21          & 28.59          & 26.55          \\
                           &                              & AdvFunMatch             & \textbf{87.49} & \textbf{63.92} & \textbf{56.08} & \textbf{54.72} & \textbf{53.37} & \textbf{66.13} & \textbf{38.83} & \textbf{35.33} & \textbf{31.54} & \textbf{29.64} \\ \bottomrule
\end{tabular}%
}
\end{table}

\subsection{Comparison with Previous Work}

As shown in Table \ref{tab:2}, we compare the performance of our AdvFunMatch method with two AT methods, PGD-AT \cite{pgd_at} and TRADES \cite{trades}, and four robustness distillation methods, ARD \cite{ard}, IAD \cite{iad}, AKD$^2$ \cite{akd}, and RSLAD \cite{rslad}, using Flip+RandomCrop augmentations. 
Implementation details for the comparison methods are provided in the Appendix A.1. 
According to the results, AdvFunMatch consistently outperforms PGD-AT, TRADES, ARD, IAD, AKD$^2$, and RSLAD in terms of clean accuracy and robust accuracy on both CIFAR-10 and CIFAR-100 datasets, considering various teacher-student model pairs such as WRN-28-10 to ResNet-18 and ResNet-18 to MobileNetV2. The improvements in clean and robust accuracy are substantial, demonstrating the effectiveness of AdvFunMatch.

\subsection{Ablation Studies}
\label{sec:4.3}
\textbf{Effect of matching on different examples}.
Table \ref{tab:3} presents a comparison of the clean and robust accuracy achieved by our AdvFunMatch framework for matching different examples, including 2-step and 10-step mismatched examples with and without using teacher gradients, 10-step adversarial examples, and 10-step RSL-derived adversarial examples. To generate these
examples, we use the same $\epsilon_{train}=8/255$ without using the additional perturbation. It can be observed that our 10-step mismatched examples
achieve the best robust accuracy of 52.53\%, followed by our 2-step mismatched examples, which achieve a robust accuracy of 51.47\%. These results significantly outperform the effects of matching other types of samples, confirming the effectiveness of the mismatched examples. However, considering that the computation time for 10-step mismatched examples is relatively long, we opt for the more efficient 2-step mismatched examples to balance the computation time and performance. In Appendix A.2, we discuss the computational efficiency of mismatched examples in greater detail and provide the computation time for different teacher-student model pairs.

\begin{table}[t]
\centering
\caption{Test accuracy (\%) of matching different examples on CIFAR-10 using ResNet-18. Best result is in \textbf{bold}.}
\label{tab:3}
\resizebox{0.4\textwidth}{!}{%
\begin{tabular}{lcccc}
\toprule
Worst case & Steps  & Clean          & PGD20          & AA          \\ \midrule
$x_{adv}$                   & 10                     & 85.42          & 52.44          & 48.99          \\
$x_{adv-RSL}$               & 10                     & 84.99          & 53.13          & 49.58          \\
$x_{mis}$                   & 10                     & 86.67          & \textbf{55.85} & \textbf{52.53} \\
$x_{mis}$                   & 2                      & \textbf{87.93} & 54.98          & 51.47          \\
$x_{mis}$ (w/o TG)          & 10                     & 84.36          & 53.25          & 49.96          \\
$x_{mis}$ (w/o TG)          & 2                      & 81.50          & 49.37          & 44.81          \\ \bottomrule
\end{tabular}%
}
\end{table}
\begin{table}[t]
  \centering
    \centering
    \caption{Test accuracy (\%) of using additional perturbation size on CIFAR-10, $\epsilon_{train}=8/255$.}
    \label{tab:4}
    \resizebox{0.3\textwidth}{!}{%
    \begin{tabular}{lccc}
    \toprule
    $\epsilon_{add}$ & Clean  & PGD20   & AA       \\
    \midrule
    0           & \textbf{87.93}  & 54.98 & 51.47 \\
    4/255                             & 86.97    & 55.75   & 52.17    \\
    6/255                             & 86.58    & 55.94   & \textbf{52.40}    \\
    8/255                             & 86.42    & \textbf{56.88}   & 52.36    \\
    \bottomrule
    \end{tabular}
    }
\end{table}

\textbf{Effect of $\epsilon_{add}$}.
Table \ref{tab:4} shows the effect of different additional perturbation sizes $\epsilon_{add}$. We can see that using an additional perturbation can further
improve robustness, with a small cost to clean accuracy. However, we also observe that the improvements brought by larger $\epsilon_{add}$ values have an upper limit. When using an $\epsilon_{add}$ of 8/255, the AA accuracy decreases by 0.04 compared to when using 6/255. Therefore, choosing an appropriate $\epsilon_{add}$ is important for balancing the robustness-accuracy tradeoff.

\subsection{Strong Data Augmentations Can Improve Robust Generalization}
Strong data augmentations have been found to have limited effectiveness in improving the robustness of AT \cite{cite8}, but they are useful in KD \cite{cons_t,wang2022makes}. This difference inspires our exploration of the impact of strong data augmentations on the performance of AdvFunMatch.
Table \ref{tab:5} presents the impact of various strong data augmentation techniques on the performance of our AdvFunMatch method, evaluating clean accuracy and robust accuracy against different attacks on CIFAR-10 and CIFAR-100 datasets, using ResNet-18 as student model. Implementation details about these augmentations can be found in Appendix A.3.
The results show that strong data augmentation enhances the performance of AdvFunMatch. For instance, CutMix demonstrates significant improvements in robust accuracy across various attacks, with gains ranging from 2.33\% to 5.62\%. Similarly, MixUp, Cutout, AutoAugment, and RandAugment also improve the performance compared to the baseline Flip+RandomCrop.

We further explore using a combination of strong augmentations, such as the integration of AutoAugment and CutMix, which provides the most considerable improvements in both clean and robust accuracy. This highlights the benefits of incorporating multiple data augmentation techniques. Overall, these findings demonstrate that employing strong data augmentation methods can further enhance the effectiveness of AdvFunMatch in defending against adversarial attacks.

Additionally, since previous robustness distillation work, such as ARD and RSLAD, have not extensively studied the effects of strong data augmentations, we further investigate their impact to better understand their potential benefits. We also examine data augmentations in Label Smoothing \cite{labelsmoothing}, a manually designed soft label method. We find that these methods can also benefit from strong augmentations, albeit not as effectively as in AdvFunMatch. This observation indicates that the limited effectiveness of strong data augmentations on robustness may be due to inappropriate label settings.
Due to space constraints, we present the results and further discuss the possible reasons for improved robustness when combining proper labels and strong data augmentations in Appendix A.4.

\begin{table}[t]
    \centering
    \caption{Effect of data augmentations in AdvFunMatch, the \textcolor{red}{red} font indicates the improvements.}
    \label{tab:5}
    \resizebox{\columnwidth}{!}{%
    \begin{tabular}{l|*{5}{c}|*{5}{c}}
    \toprule
    \multirow{2}{*}{Data Augmentation} & \multicolumn{5}{c|}{CIFAR-10}          & \multicolumn{5}{c}{CIFAR-100}          \\
                                       & Clean & FGSM  & PGD20 & CW${\infty}$     & AA    & Clean & FGSM  & PGD20 & CW${\infty}$     & AA    \\ \midrule
    Flip+RandomCrop                    & 86.58 & 64.28 & 55.94  & 54.80 & 52.40 & 64.30 & 38.36 & 34.29  & 30.65 & 28.80  \\
    MixUp                              & 88.35/\textcolor{red}{+1.77} & 66.10/\textcolor{red}{+1.82} & 59.06/\textcolor{red}{+3.12}  & 57.25/\textcolor{red}{+2.45} & 54.98/\textcolor{red}{+2.58} & 67.02/\textcolor{red}{+2.72} & 40.52/\textcolor{red}{+2.16} & 36.36/\textcolor{red}{+2.07}  & 31.92/\textcolor{red}{+1.27} & 30.26/\textcolor{red}{+1.46}      \\
    Cutout                             & 88.17/\textcolor{red}{+1.59} & 65.81/\textcolor{red}{+1.53} & 58.01/\textcolor{red}{+2.07}  & 56.19/\textcolor{red}{+1.39} & 53.62/\textcolor{red}{+1.22} & 65.13/\textcolor{red}{+0.83} & 39.53/\textcolor{red}{+1.17} & 35.43/\textcolor{red}{+1.14}  & 31.26/\textcolor{red}{+0.61} & 29.67/\textcolor{red}{+0.87}       \\
    CutMix                             & 87.71/\textcolor{red}{+1.13} & 66.61/\textcolor{red}{+2.33} & 61.56/\textcolor{red}{+5.62}  & 58.71/\textcolor{red}{+3.91} & 57.02/\textcolor{red}{+4.62} & 64.92/\textcolor{red}{+0.62} & 40.36/\textcolor{red}{+2.00} & 37.07/\textcolor{red}{+2.78}  & 32.35/\textcolor{red}{+1.70} & 30.98/\textcolor{red}{+2.18} \\
    AutoAugment                        & 88.58/\textcolor{red}{+2.00} & 66.09/\textcolor{red}{+1.81} & 58.97/\textcolor{red}{+3.03}  & 57.45/\textcolor{red}{+2.65} & 55.40/\textcolor{red}{+3.00} & 66.41/\textcolor{red}{+2.11} & 39.43/\textcolor{red}{+1.07} & 35.56/\textcolor{red}{+1.27}  & 31.11/\textcolor{red}{+0.46} & 29.49/\textcolor{red}{+0.69} \\
    RandAugment                         & 88.82/\textcolor{red}{+2.24} & 66.01/\textcolor{red}{+1.73} & 58.56/\textcolor{red}{+2.62}  & 56.75/\textcolor{red}{+1.95} & 54.51/\textcolor{red}{+2.11} & 66.91/\textcolor{red}{+2.61} & 39.42/\textcolor{red}{+1.06} & 35.11/\textcolor{red}{+0.82}  & 31.21/\textcolor{red}{+0.56} & 29.44/\textcolor{red}{+0.64} \\
      AutoAugment+CutMix                 & 88.83/\textcolor{red}{+2.25} & 67.72/\textcolor{red}{+3.44} & 61.01/\textcolor{red}{+5.07}  & 58.62/\textcolor{red}{+3.82} & 56.71/\textcolor{red}{+4.31} & 66.65/\textcolor{red}{+2.35} & 40.56/\textcolor{red}{+2.20} & 37.23/\textcolor{red}{+2.94}  & 32.24/\textcolor{red}{+1.59} & 30.67/\textcolor{red}{+1.87}       \\ \bottomrule

    \end{tabular}
    }
\end{table}

\subsection{Longer Training Epochs Lead to Improved Performance}
\begin{figure*}[t]
\vskip 0.2in
\center{
    \begin{subfigure}{0.24
    \linewidth}
        \centering
        \includegraphics[width=\textwidth]{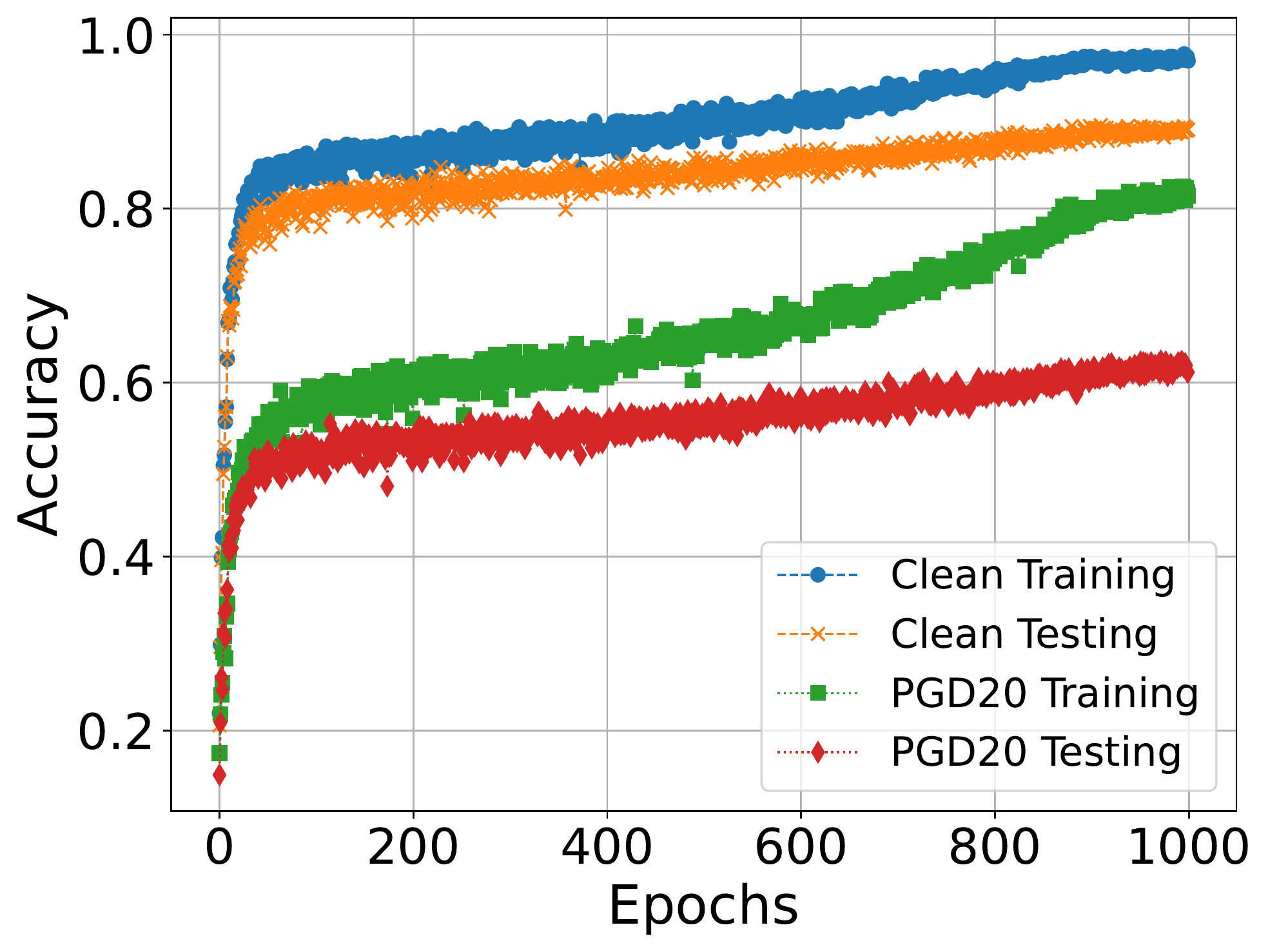}
        \caption{ResNet-18,CIFAR-10}
    \end{subfigure}
    \begin{subfigure}{0.24\linewidth}
        \centering
        \includegraphics[width=\textwidth]{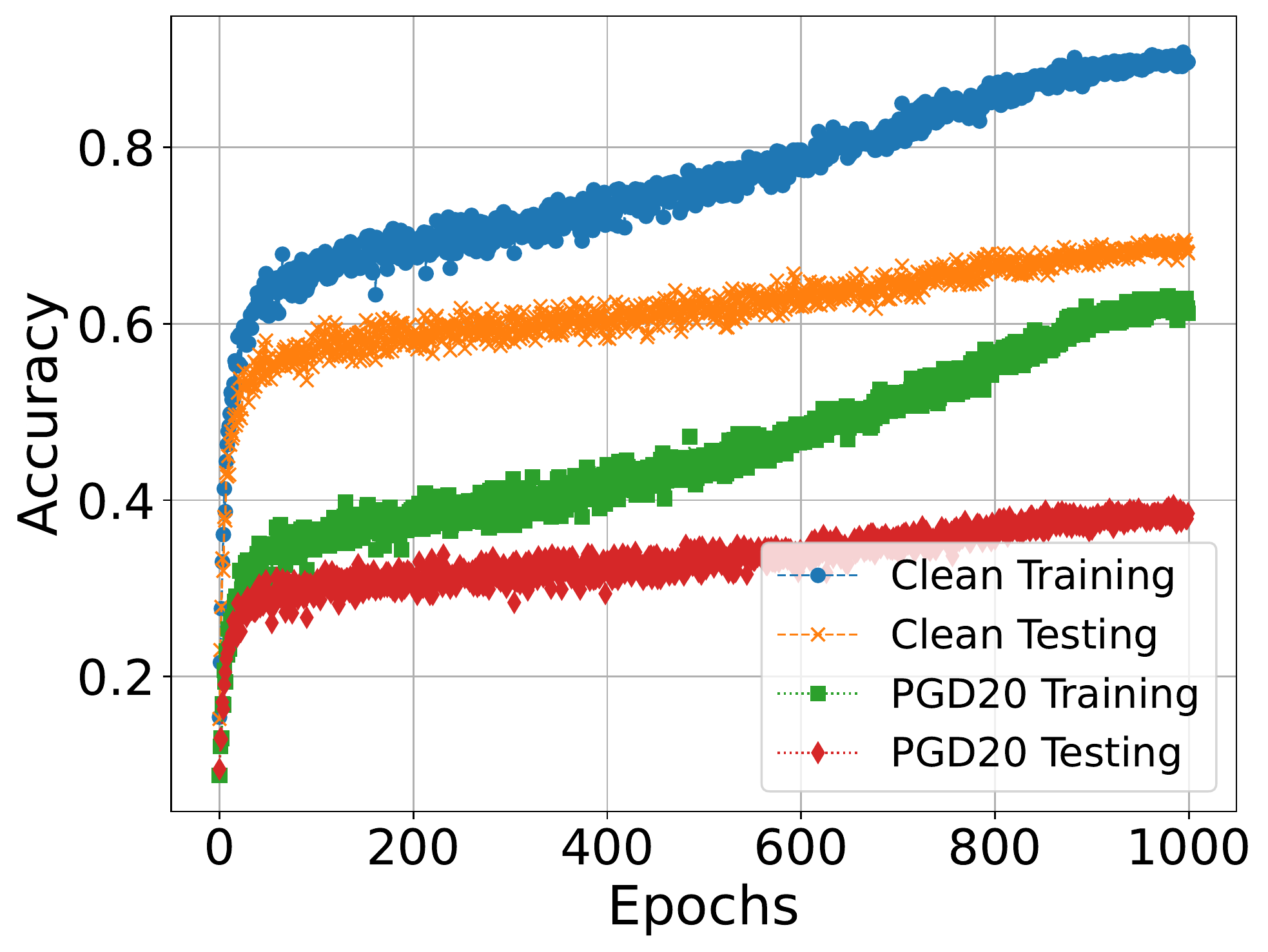}
        \caption{ResNet-18,CIFAR-100}
    \end{subfigure}
    \begin{subfigure}{0.24\linewidth}
        \centering
        \includegraphics[width=\textwidth]{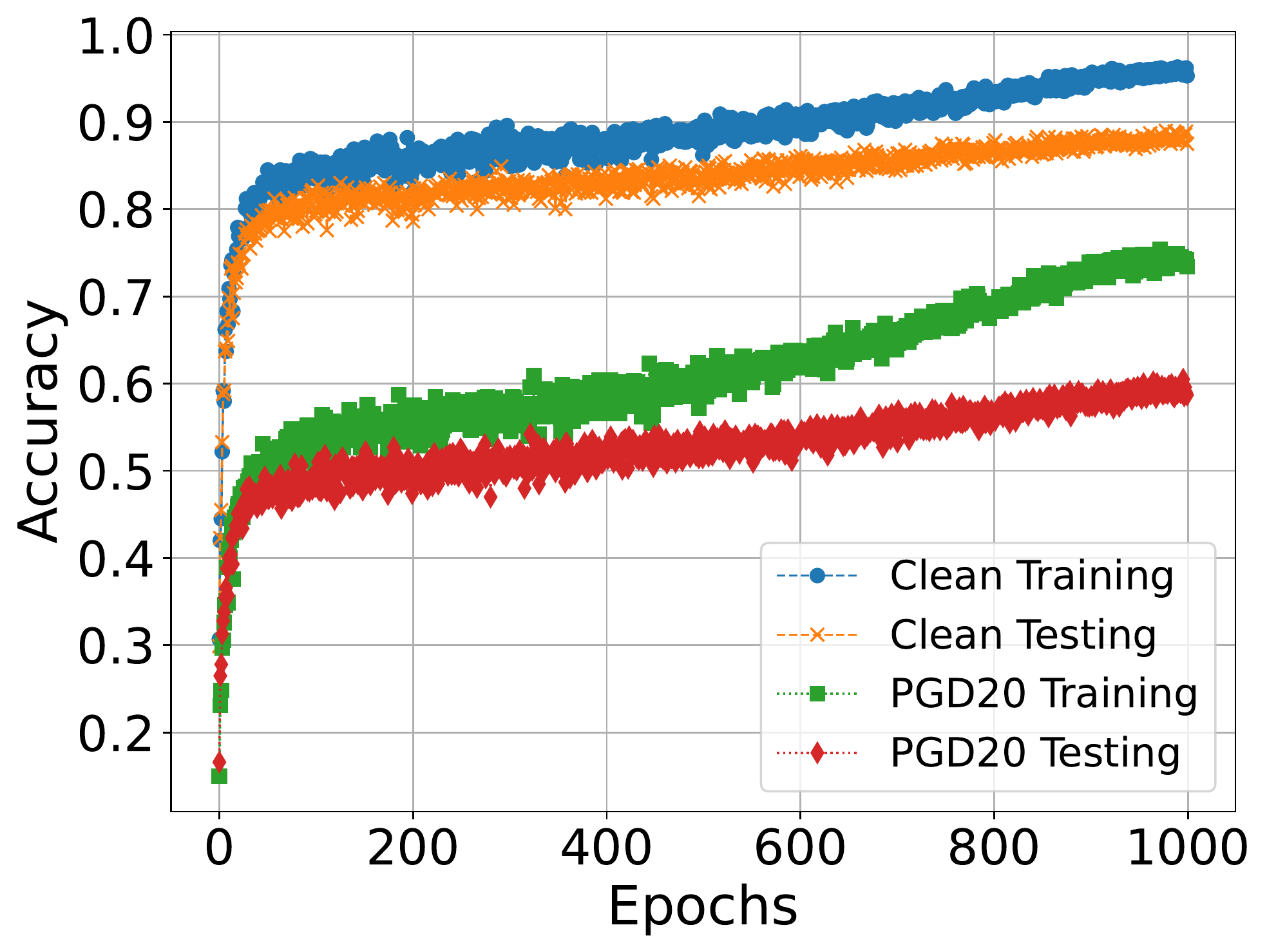}
        \caption{MNV2,CIFAR-10}
    \end{subfigure}
        \begin{subfigure}{0.24\linewidth}
        \centering
        \includegraphics[width=\textwidth]{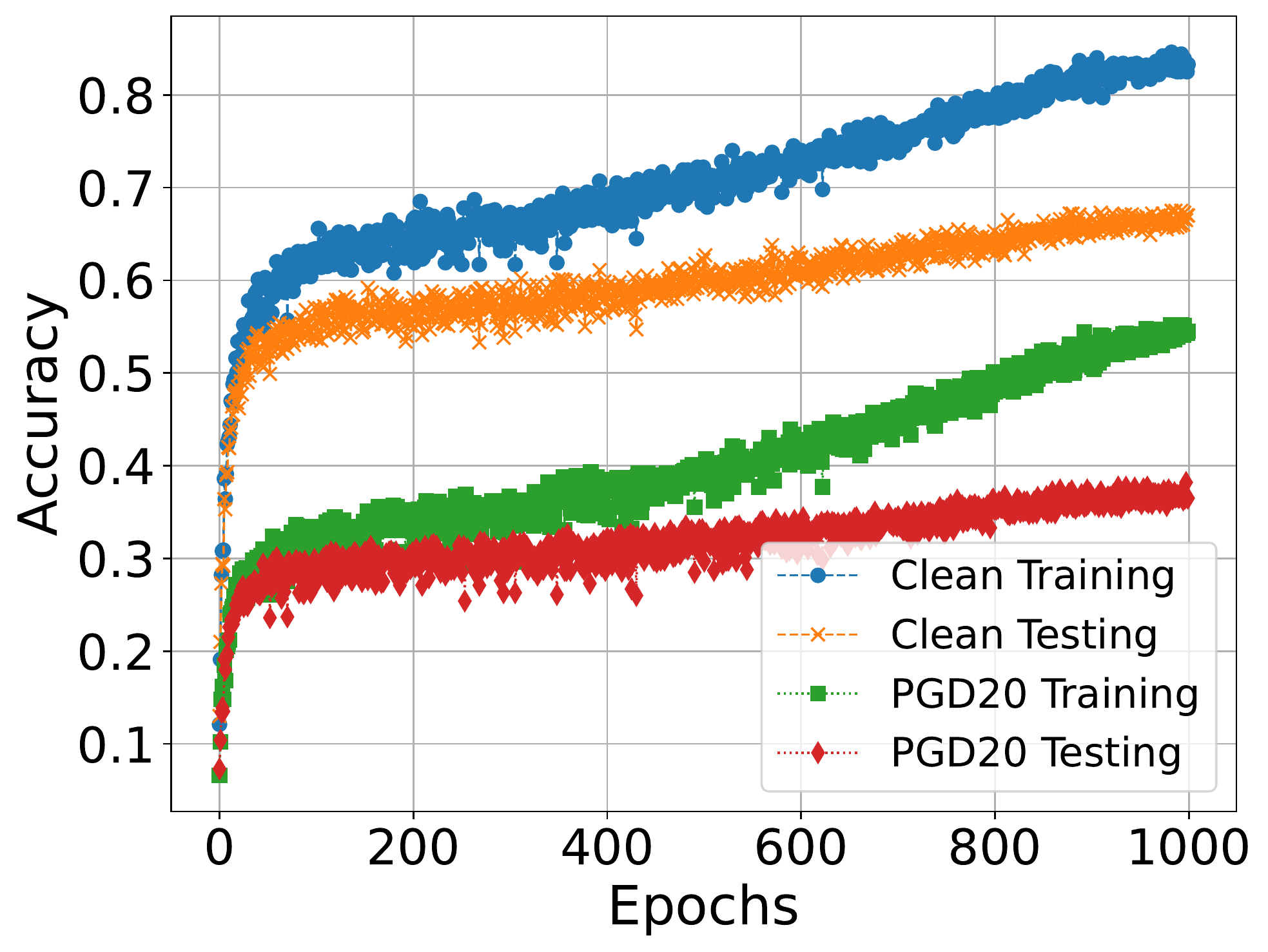}
        \caption{MNV2,CIFAR-100}
    \end{subfigure}
}
    \caption{1000-epoch training and testing accuracy curve, MNV2 denotes MobileNetV2.}
\label{fig:5}
\vskip -0.2in
\end{figure*}
\begin{table}[ht]
    \centering
    \caption{The SOTA robust accuracy (\%) after 1000-epoch training.}
    \label{tab:6}
    \resizebox{0.5\textwidth}{!}{%
    \begin{tabular}{llccccc}
    \toprule
    Model                        & Dataset   & Clean & FGSM  & PGD20 & CW${\infty}$ & AA    \\ \midrule
    \multirow{2}{*}{ResNet-18}   & CIFAR-10  & 89.08   & 69.47 & 62.53 & 60.26        & 58.30 \\
                                 & CIFAR-100 & 67.55   & 41.48 & 38.19 & 33.63        & 32.37 \\
    \multirow{2}{*}{MobileNetV2} & CIFAR-10  & 87.82   & 65.56 & 59.92 & 57.32        & 55.70 \\
                                 & CIFAR-100 & 66.16   & 39.87 & 37.09 & 32.74        & 30.95 \\ \midrule
    \end{tabular}%
    }
\end{table}

As suggested by \cite{cons_t,Stanton}, extending the training epochs can aid the student in better matching the teacher. By increasing the training epochs to 1000 using the AutoAugment+CutMix augmentations, we observe improvements in both clean and robust accuracy compared to the 300-epoch training, as demonstrated in Table \ref{tab:6}. Remarkably, these results achieve a new SOTA on the RobustBench Leaderboard \cite{robustbench} for the same model architecture and even exhibit greater robustness than some larger WRN models. Additionally, we investigate the possibility of robust overfitting through the accuracy curve, illustrated in Figure \ref{fig:5}, and find that there is no overfitting even at 1000 training epochs.

\section{Conclusion}
\label{sec:5}

In this paper, we presented AdvFunMatch, a novel robustness distillation method that transfers adversarial robustness from a teacher model to a student model using consistent teaching principles. We introduced the concept of mismatched examples, generated through teacher gradient guidance and an aggressive perturbation strategy. By aligning the distributions of teacher and student models within the norm ball of augmented samples, AdvFunMatch effectively improves clean and robust accuracy on benchmark datasets, surpassing previous methods. Our work contributes to the field of robustness distillation by proposing a simple yet effective framework based on consistent teaching and mismatched examples.

\textbf{Limitation}. One limitation of our work is its reliance on gradients from a differentiable teacher model to guide the generation of mismatched examples. In real-world scenarios, however, the teacher model may not always be differentiable, such as when only an API is provided. Distilling models in such scenarios is also known as model stealing attacks \cite{tramer2016stealing}. Therefore, future research should focus on developing mismatched example optimization methods that do not depend on teacher gradients, enabling the application of AdvFunMatch to teacher models in black-box scenarios.

{
\small
    \bibliographystyle{abbrv}
    \bibliography{adv_training}

}



\appendix

\section{Appendix}
\subsection{Details of Comparison Methods}
In this paper, we compare our AdvFunMatch with two Adversarial Training (AT) methods, PGD-AT and TRADES, and four robustness distillation methods, ARD, IAD, AKD2, and RSLAD.

We adopt the implementation of PGD-AT and TRADES from Rice et al. [25], while for ARD, IAD, AKD2, and RSLAD, we utilize the code from their respective Github repositories \footnote{ARD:\url{https://github.com/goldblum/AdversariallyRobustDistillation}, IAD:\url{https://github.com/ZFancy/IAD}, RSLAD:\url{https://github.com/zibojia/RSLAD}, AKD$^{2}$:\url{https://github.com/VITA-Group/Alleviate-Robust-Overfitting}} and follow their settings. Specifically, for IAD, we implement their IAD-I variant. Moreover, AKD$^2$ and IAD utilize self-distillation in their experiments, while we substitute the same WRN-28-10 and ResNet-18 teacher models as used in our AdvFunMatch for a fair comparison.

\subsection{Computational Efficiency}
In this section, we discuss the computational efficiency of generating our 2-step mismatched examples. Although the number of PGD iterations is reduced compared to 10-step adversarial examples, it is not straightforward to conclude that 2-step mismatched examples are more computationally efficient. This is because we utilize both teacher and student gradients to generate mismatched examples, which may affect the overall efficiency.

By using FB${_t}$ and FB${_s}$ to denote the one-time forward-backward pass of the teacher model and the student model, the generation time of our 2-step mismatched examples is 2FB${_t}$ + 2FB${_s}$, and the 10-step adversarial examples is 10FB${_s}$. The difference is 8FB${_s}$ - 2FB$_{_t}$, which means that when FB$_t$ < 4FB$_s$, our 2-step mismatched examples are more efficient to generate than the 10-step adversarial examples generated from the student model. Therefore, choosing appropriate teacher-student pairs can lead to faster training. For instance, with a ResNet-18 teacher and MobileNetV2 student models, our 2-step
mismatched examples are about 2.8× faster than 10-step adversarial examples, as shown in Table \ref{tab:8}.

When the computational expense of the teacher model significantly exceeds that of the student model, such that FB$_t$ $\geq$ 4FB$_s$ (as is the case with the WRN-28-10 teacher and ResNet-18 student), generating 2-step mismatched examples can be slower than generating 10-step adversarial examples. However, even in such cases, the generation time for our 2-step mismatched example typically falls within 20-25\% of the time required to generate 10-step adversarial examples using the larger teacher model. This accounts for 20\% when the teacher model is substantially larger than the student model (FB$_t$ $\gg$ 4FB$_s$), and 25\% when FB$_t$ = 4FB$_s$.

To validate this hypothesis, we tested the time taken to generate 10-step adversarial examples using WRN-28-10, which resulted in 399 seconds/epoch. In contrast, under the WRN-28-10-ResNet-18 configuration, generating 2-step mismatched examples took approximately 94 seconds/epoch, which is about 23.55\% of the former.

Consequently, our method consistently trains student models with significantly lower computational overhead compared to training larger, robust teacher models. This makes it a practical choice for real-world applications.

\setcounter{table}{6}
\begin{table}[ht]
  \centering
    \caption{Generation time for different types of examples, calculated using an RTX A5000 GPU.}
    \label{tab:8}
    \resizebox{0.5\textwidth}{!}{%
    \begin{tabular}{lccccc}
    \toprule
    \multirow{2}{*}{Teacher} & \multirow{2}{*}{Student}  & \multicolumn{3}{c}{Generation time/Epoch (sec)}\\ \cmidrule(lr){3-5}
                            &     & $x_{adv}$  & $x_{adv-RSL}$    &  $x_{mis}$   & \\ \midrule
                  & WRN-28-10                                       & 399     & -   & -  \\
    WRN-28-10            & ResNet-18                                       & 56     & 70   & 94  \\
    ResNet-18         &  MobileNetV2                             & 98    & 100   & 35 & \\

    \bottomrule
    \end{tabular}
    }

\end{table}

\subsection{Detials of Augmentations}
\textbf{MixUp and CutMix}.
MixUp generates a new sample by linearly interpolating the input images and their corresponding labels. Given two images $x_1$ and $x_2$ with their labels $y_1$ and $y_2$, MixUp creates a new image $x_{mix}$ and its label $y_{mix}$ using a mixing parameter $\lambda \in [0, 1]$.
CutMix is another data augmentation technique that combines two images by replacing a randomly selected region of one image with the corresponding region from another image. The label of the new image is also a mixture of the original labels, weighted by the area of the replaced region. In this paper, we
do not use the interpolation label, but only take advantage of interpolation image for distillation. For both MixUp and CutMix, we sample the mixing parameter
$\lambda$ from a beta distribution $\beta(\alpha,\alpha)$ with $\alpha=1$.

\textbf{Cutout}.
Cutout works by randomly removing (zeroing out) a rectangular region within an input image during training. This encourages the model to learn more contextual information and reduces its reliance on local features. For CIFAR-10, we set the length of the randomly removed patch to 16, and for CIFAR-100, the length is set to 8.

\textbf{AutoAugment}.
AutoAugment is a data augmentation technique that aims to improve the performance of deep learning models by automatically searching for the optimal augmentation policies for a given dataset. We use the searched CIFAR policy in this paper.

\textbf{RandAugment}.
RandAugment is a data augmentation technique designed to improve the performance of deep learning models by applying a random set of image transformations to the training dataset. It simplifies the search for optimal augmentation policies compared to AutoAugment. Instead of using reinforcement learning to find the best combination of augmentations, RandAugment only requires two hyperparameters: the number of transformations to apply (N) and their magnitude (M).
In this paper, we set N=1 and M=2 for both CIFAR-10/CIFAR-100 datasets.

\textbf{Combination of augmentations}.
We also study the effect of combining different strong augmentations, e.g., AutoAugment + CutMix. To avoid affecting the performance of the original augmentations, we use a random manner to decide whether to apply AutoAugment or CutMix to the current training sample, with the random probability $p$ set to 0.5.

\subsection{Effect of Data Augmentation in ARD/RSLAD/LS}
\begin{table}[h]
\centering
\caption{Test accuracy (\%) of using AutoAugment in ARD/RSLAD/LS on CIFAR-10.}
\label{tab:7}
\resizebox{0.4\textwidth}{!}{%
\begin{tabular}{lccc}
\toprule
Method       & Clean & PGD20 & AA    \\ \midrule
PGD-AT       & 81.58 & 49.96 & 46.25 \\
+AutoAugment & 85.01 & 50.81 & 46.42 \\
Diff &  +3.43 & +0.85 & +0.17 \\ \midrule
PGD-AT+LS    & 82.27 & 51.19 & 46.99 \\
+AutoAugment & 86.56 & 52.32 & 47.43 \\
Diff  & +4.29 & +1.13 & +0.44 \\ \midrule
ARD          & 83.37 & 52.80 & 49.18 \\
+AutoAugment & 84.68 & 54.73 & 50.50 \\
Diff & +1.31 & +1.93 & +1.32 \\ \midrule
RSLAD        & 84.91 & 52.89 & 49.03 \\
+AutoAugment & 87.28 & 56.36 & 52.29 \\
Diff & +2.37 & +3.47 & +3.26\\ \bottomrule
\end{tabular}%
}
\end{table}

In Table \ref{tab:7}, we present the results of PGD-AT, PGD-AT+Label Smoothing(LS), ARD, and RSLAD using Flip+RandomCrop and AutoAugment. For LS, we set
the smoothing parameter $\alpha_{ls}=0.5$.
We can observe that AutoAugment only marginally improves the AA accuracy of PGD-AT by 0.17. When using LS, the improvement of AutoAugment is more significant at 0.44. For ARD and RSLAD, employing AutoAugment results in increases of 1.32 and 3.27 in AA accuracy, respectively. These observations indicate that the limited effectiveness of strong data augmentations on robustness may be due to inappropriate label settings.

We offer two conjectures to help understanding the improvement in robustness when combining data augmentation and soft labels.

\textbf{Conjecture 1}. From the function matching perspective, matching on any input should advance the student model's progress in aligning with the teacher model. As stronger augmentations improve the input diversity,
matching on samples with strong data augmentations can help the student model better align with the robust teacher model, thereby enhancing its robustness. This perspective can help understanding the significant robustness improvements observed when using data augmentations in our AdvFunMatch.

\textbf{Conjecture 2}. From the domain generalization perspective, stronger data augmentations can result in larger domain shifts between the augmented data distribution and the test data distribution. In such cases, training with hard labels can lead to over-confident robust models on the augmented data, which do not guarantee good robust generalization on the test data. By using appropriate soft labels for training on augmented data, the domain shift can be reduced, thereby improving robustness. This perspective could help understanding why LS/ARD/RSLAD can also improve the effectiveness of strong data augmentations in enhancing robustness.

\end{document}